\documentclass[sigconf, noacm, dvipsnames]{acmart}

\AtBeginDocument{%
  \providecommand\BibTeX{{%
    \normalfont B\kern-0.5em{\scshape i\kern-0.25em b}\kern-0.8em\TeX}}}

\newcommand\ignore[1]{ }
\usepackage[utf8]{inputenc}
\usepackage{enumitem} 
\usepackage{xcolor}
\usepackage{multirow}
\usepackage[colorinlistoftodos,prependcaption,textsize=small]{todonotes}
\usepackage{cleveref}
\DeclareUnicodeCharacter{2212}{\ensuremath{-}}
\crefformat{section}{\S#2#1#3}
\crefformat{subsection}{\S#2#1#3}
\crefformat{subsubsection}{\S#2#1#3}
  \usepackage{array}
\definecolor{airforceblue}{rgb}{0.36, 0.54, 0.66}
\definecolor{dodgerblue}{rgb}{0.12, 0.56, 1.0}
\definecolor{brandeisblue}{rgb}{0.0, 0.44, 1.0}
\definecolor{brickred}{rgb}{0.8, 0.25, 0.33}
\definecolor{eggplant}{rgb}{0.38, 0.25, 0.32}
\definecolor{byzantium}{rgb}{0.44, 0.16, 0.39}
\definecolor{dgreen}{rgb}{0.00, 0.75, 0.00}

\usepackage{tikz}
\usepackage{booktabs}
\usepackage{tabularx}
\usepackage{siunitx}
\usepackage{caption}
\definecolor{headercolor}{RGB}{210, 235, 255}
\definecolor{rowcolor}{RGB}{240, 240, 240}
\usepackage{newunicodechar}
\newcommand*\circled[1]{\tikz[baseline=(char.base)]{
            \node[shape=circle,draw,inner sep=1pt] (char) {#1};}}

\def\BibTeX{{\rm B\kern-.05em{\sc i\kern-.025em b}\kern-.08em
    T\kern-.1667em\lower.7ex\hbox{E}\kern-.125emX}}

\usepackage{anyfontsize}
\newunicodechar{≤}{\ensuremath{\leq}}
\definecolor{brickred}{rgb}{0.8, 0.25, 0.33}
\definecolor{dblue}{rgb}{0.1, 0.2, 0.70}
\newcommand{\ms}[1]{\textcolor{black}{#1}}

\newcommand{\kailash}[1]{\textcolor{black}{#1}}

\newcommand{\name}{{SwiftRL}\xspace}
\usepackage{tikz}
\usetikzlibrary{shapes.geometric, arrows, positioning, shadows}

\pgfdeclarelayer{background}
\pgfsetlayers{background,main}

\tikzstyle{texte} = [above]

\tikzstyle{stop} = [rectangle, rounded corners, minimum width=3cm, minimum height=1cm,text centered, draw=black, fill=violet!30, drop shadow]
\tikzstyle{io} = [trapezium, trapezium left angle=70, trapezium right angle=110, text centered, text width=5cm, draw=black, fill=blue!30, drop shadow]
\tikzstyle{processcpu} = [rectangle, text centered, text width=5cm, draw=black, fill=orange!30, drop shadow]
\tikzstyle{processdpu} = [rectangle, text centered, text width=5cm, draw=black, fill=red!30, drop shadow]
\tikzstyle{decision} = [diamond, aspect=2, text centered, draw=black, fill=green!30, drop shadow]
\tikzstyle{arrow} = [thick,->,>=stealth]

\usepackage{algorithm}
\usepackage{algpseudocode}
\usepackage{stmaryrd}
\usepackage{amsmath}
\usepackage{listings}
\usepackage{subcaption}

\definecolor{mygreen}{rgb}{0,0.6,0}
\definecolor{mygray}{rgb}{0.5,0.5,0.5}
\definecolor{mymauve}{rgb}{0.58,0,0.82}

\lstdefinestyle{myC}{
  language=Matlab,
  backgroundcolor=\color{white},  
  basicstyle=\footnotesize,        
  breakatwhitespace=false,  
  breaklines=true,       
  captionpos=b,                   
  commentstyle=\color{mygreen},    
  deletekeywords={...},           
  escapechar=\%,
  xleftmargin=0pt,
  xrightmargin=0pt,
  aboveskip=\medskipamount,
  belowskip=\medskipamount,
  extendedchars=true,            
  keepspaces=true,               
  keywordstyle=\color{blue},      
  language=C++,              
  morekeywords={__builtin_mul_sl_ul_rrr,__builtin_mul_sl_sh_rrr,mul_ul_ul,mul_sh_ul,mul_sh_sh,mul_sl_ul,mul_sl_sh,lbs,lhs,move,lsl_add,lw,add,sw,jneq,mem_alloc, *,...},          
  numbers=left,                   
  numbersep=1pt,                   
  numberstyle=\tiny\color{mygray}, 
  rulecolor=\color{black},     
  showspaces=false,                
  showstringspaces=false,          
  showtabs=false,                  
  stepnumber=1,                    
  stringstyle=\color{mymauve},     
  tabsize=2,	                   
  title=\lstname                   
}

\definecolor{bluehl}{rgb}{0.8,0.874,1}
\definecolor{pinkhl}{rgb}{0.992156863,0.847058824,1}
\definecolor{macaroniandcheese}{rgb}{1.0, 0.74, 0.53}
\definecolor{mossgreen}{rgb}{0.68, 0.87, 0.68}
\definecolor{greenhl}{rgb}{0.835,0.996,0.939}
\definecolor{yellowhl}{rgb}{0.996,0.957,0.8}
\definecolor{palecerulean}{rgb}{0.61, 0.77, 0.89}
\definecolor{gray(x11gray)}{rgb}{0.75, 0.75, 0.75}

\AtBeginDocument{\DeclareCaptionSubType{lstlisting}}

\setcopyright{none}
\renewcommand\footnotetextcopyrightpermission[1]{}

\makeatletter
\def\bstctlcite{\@ifnextchar[{\@bstctlcite}{\@bstctlcite[@auxout]}}
\def\@bstctlcite[#1]#2{\@bsphack
  \@for\@citeb:=#2\do{%
    \edef\@citeb{\expandafter\@firstofone\@citeb}%
    \if@filesw\immediate\write\csname #1\endcsname{\string\citation{\@citeb}}\fi}%
  \@esphack}
\makeatother

\usepackage{xspace}

\ignore{
\newcommandx{\unsure}[2][1=]{\todo[linecolor=red,backgroundcolor=red!25,bordercolor=red,#1, size=\tiny]{#2}}
\newcommandx{\feedback}[2][1=]{\todo[linecolor=yellow,backgroundcolor=yellow!25,bordercolor=yellow,#1]{#2}}

\titlespacing\section{2pt}{3pt plus 1pt minus 1pt}{2pt plus 1pt minus 1pt}
\titlespacing\subsection{2pt}{3pt plus 1pt minus 1pt}{2pt plus 1pt minus 1pt}
\titlespacing\subsubsection{2pt}{3pt plus 1pt minus 1pt}{2pt plus 1pt minus 1pt}

\makeatletter
\g@addto@macro{\normalsize}{%
  \setlength{\abovedisplayskip}{2pt plus 1pt minus 1pt}
  \setlength{\belowdisplayskip}{2pt plus 1pt minus 1pt}
  \setlength{\abovedisplayshortskip}{0pt}
  \setlength{\belowdisplayshortskip}{0pt}
  \setlength{\intextsep}{2pt plus 1pt minus 1pt}
  \setlength{\textfloatsep}{3pt plus 1pt minus 1pt}
  \setlength{\dbltextfloatsep}{3pt plus 1pt minus 1pt}
  \setlength{\skip\footins}{4pt plus 1pt minus 1pt}}
  \setlength{\abovecaptionskip}{2pt plus 1pt minus 1pt}
\makeatother
}

\newcounter{take}
\newcounter{reco}
\usepackage[T1]{fontenc}
\usepackage[utf8]{inputenc}
\newcommand{\tsc}[1]{\textsuperscript{#1}} 
\newcommand{\affilGWU}{\tsc{1}}
\newcommand{\affilETH}{\tsc{2}}
\newcommand{\affilIND}{\tsc{3}}

\begin{document}
\bstctlcite{IEEEexample:BSTcontrol}

\title{
SwiftRL:~Towards Efficient Reinforcement Learning on Real Processing-In-Memory Systems}

\author{
 {%
     Kailash Gogineni$^1$\quad 
     Sai Santosh Dayapule$^1$\quad 
     Juan Gómez-Luna$^2$\quad 
     Karthikeya Gogineni$^3$
 }
}
\author{
 {
     Peng Wei$^1$\quad
     Tian Lan$^1$\quad
     Mohammad Sadrosadati$^2$\quad
     Onur Mutlu$^2$\quad
     Guru Venkataramani$^1$
 }
}


\affiliation{
\institution{
      \vspace{5pt}
      \affilGWU George Washington University, USA \quad
      \affilETH ETH Zürich, Switzerland \quad
      \affilIND Independent\quad
  }
}

\pagestyle{plain}

\begin{abstract}
\kailash{Reinforcement Learning (RL) is the process by which an agent learns optimal behavior through interactions with experience datasets, all of which aim to maximize the reward signal.} RL algorithms often face performance challenges in real-world applications, especially when training with extensive and diverse datasets. For instance, applications like autonomous vehicles include sensory data, dynamic traffic information (\kailash{including movements} of other vehicles and pedestrians), critical risk assessments, and varied agent actions. Consequently, RL training is significantly memory-bound due to sampling large experience datasets that may not fit entirely into the \kailash{hardware caches} and frequent data transfers needed between \kailash{memory and the computation units} (e.g., CPU, GPU), especially during batch updates. This bottleneck results in significant execution latencies and impacts the overall training time. To alleviate such issues, recently proposed memory-centric computing paradigms, like Processing-In-Memory (PIM), can \kailash{address memory} latency-related bottlenecks by performing the computations inside the memory devices.

In this paper, we present \name, which explores the potential of real-world PIM architectures to accelerate popular RL workloads and their training \kailash{phases}. We adapt RL algorithms, namely Tabular Q-learning and SARSA, on UPMEM PIM systems and first observe their performance using two different environments and three sampling strategies. We then implement performance optimization strategies during RL adaptation to PIM by approximating the Q-value update function \kailash{(which avoids high performance costs due to runtime instruction emulation used by runtime libraries)} and incorporating certain PIM-specific routines specifically needed by the underlying algorithms. Moreover, we develop and assess a multi-agent version of Q-learning optimized for hardware and illustrate how PIM can be leveraged for algorithmic scaling with multiple agents. We experimentally \kailash{evaluate RL workloads} on OpenAI GYM environments using UPMEM hardware. Our results demonstrate a near-linear scaling of 15$\times$ in performance when the number of PIM cores increases by 16$\times$ (125 to 2000). We also compare our PIM implementation against Intel(R) Xeon(R) Silver 4110 CPU and NVIDIA RTX~3090 GPU and observe superior performance \kailash{on the UPMEM PIM System for different implementations.}

\end{abstract}

\keywords{Reinforcement learning, Processing-in-memory, \kailash{Multi-agent systems}, Memory bottleneck, Performance analysis}

\maketitle

\section{Introduction}
\label{Introduction}

\kailash{In recent years, Reinforcement Learning~(RL) has seen important breakthroughs in various domains such as robotics, games, and healthcare~\cite{sutton2018reinforcement, vinyals2019grandmaster, akkaya2019solving, silver2016mastering, yu2021reinforcement}}. All of these applications involve active interactions with the environment, from \kailash{which observations} are made in order to train the RL agent. Extending RL to real-world applications presents challenges, particularly in scenarios such as self-driving cars, where exploration and training in the field can be impractical and may even raise safety concerns while piloting a car due to delayed decisions stemming from the performance bottlenecks of underlying RL-based decision-making modules~\cite{prudencio2023survey, levine2020offline}.

Learning effective RL policies using \ms{\emph{pre-collected}} experience datasets reduces safety risks and the need for real-time interactions with the environment during training~\cite{agarwal2020optimistic, levine2020offline, prudencio2023survey}. In this setting, a behavior policy interacts with the environment to collect a set of experiences and learns the optimal policy \kailash{from pre-generated datasets during the training phase. Such offline RL has achieved} considerable success in a diverse set of safety-critical applications, including healthcare decision-making, robotic manipulation skills, and certain recommendation systems~\cite{levine2020offline, tang2022leveraging, zhan2022deepthermal}. Nevertheless, training from logs to learn a behavior policy and making \textit{data-driven decisions} is a performance-intensive process, and there may be a vast amount of data points during the training phase~\cite{levine2020offline}. Furthermore, frequent (re)training will be necessary on newly acquired data on such safety-critical applications, where modern processor-centric systems face the challenge of having to \kailash{perform costly data movement between memory and processor units before performing RL computations, negatively impacting both the total execution time and the resulting \kailash{energy consumption~\cite{mutlu2019processing, mutlu2022modern, boroumand2018google, boroumand2021google}}}.

The \kailash{Processing-In-Memory~(PIM)~\cite{mutlu2019processing, mutlu2022modern, ghose2019processing, seshadri2019dram, mutlu2019enabling}} computing paradigm, which places the processing elements inside or close to the memory chips, is well positioned to address the performance bottlenecks of memory-intensive \kailash{workloads}. Despite being researched for decades, real-world PIM chips have only recently entered the commercial market. The UPMEM PIM computing \kailash{platform~\cite{devaux2019true}} is the first commercially available architecture designed to accelerate memory-bound workloads~\cite{UPMEM, UPMEM_1, gomez2021benchmarking, gomez2022benchmarking, gomez2022machine, gomez2023evaluating, giannoula2022sparsep}. \kailash{Recent studies leverage PIM architectures to provide high performance and energy benefits on \ms{bioinformatics}, neural networks, machine learning, database kernels, homomorphic operations and more~\cite{gomez2022benchmarking, gomez2021benchmarking, giannoula2022towards, oliveira2023transpimlib, chen2023simplepim, abecassis2023gapim, diab2022high,rhyner2024analysis, giannoula2024accelerating, giannoula2022sparsep, diab2023framework,gupta2023evaluating, gomez2022machine, gomez2023evaluating,oliveira2022accelerating, das2022implementation,lim2023design, chen2023uppipe,wu2023pim}}. However, no prior work has explored the adaptation of RL workloads on \kailash{this real-world} PIM architecture and evaluated its potential to accelerate the RL training phase, \kailash{which is critical in efficiently learning effective policies.}

\kailash{In this paper, we present \name, \kailash{where we accelerate  RL algorithms}, namely Tabular Q-learning~\cite{sutton2018reinforcement, watkins1992q, li2020sample} and SARSA~\cite{sutton2018reinforcement}, on UPMEM PIM systems and measure their performance using two different environments and three sampling strategies. We implement performance optimization strategies during RL adaptation to \ms{the} PIM system via approximating the Q-value update function \kailash{(which avoids high performance costs due to emulation \ms{used} by runtime libraries)} and by adding certain custom PIM-specific routines needed by the underlying algorithms. Further, we evaluate the multi-agent version of Q-learning, showing how a real PIM \ms{system} may be used for algorithmic scaling with multiple agents.} Our experimental analysis demonstrates the performance and scalability of RL workloads across thousands of PIM cores on real-world OpenAI environments~\cite{brockman2016openai}.

In summary, our paper makes the following contributions:
\begin{itemize}
    \item We present a roofline model that highlights the memory-bounded behavior of RL workloads during \kailash{their training phases}. This motivates our  \name design that accelerates the RL algorithms with PIM cores attached to the memory banks responsible for storing training datasets.
    \item We study the benefit of \kailash{real} in-memory computing systems on two RL algorithms learning under two distinct environments and various sampling strategies for experience data: sequential, stride-based, and random.
    \item We conduct scalability~(strong scaling) tests by evaluating our RL workloads on thousands of PIM cores. Across all of our workloads, we observe a \ms{near-linear} scaling of $15\times$ in performance when the number of cores increases by $16\times$ (125 to 2000 PIM cores).
    \item \kailash{Our experimental results demonstrate superior performance of \ms{the} real PIM system over \ms{implementations} on Intel(R) Xeon(R) Silver 4110 CPU and NVIDIA RTX~3090 GPU, where the measured performance speedups of PIM adaptations are {\it at least} $1.62\times$ and $4.84\times$ respectively.
    \item We open-source our PIM implementations of RL training workloads at \textcolor{black}{\url{https://github.com/kailashg26/SwiftRL}}}.
\end{itemize}

\section{Background and Motivation}
\label{Motivation}

\subsection{Reinforcement Learning}

\kailash{Reinforcement learning is a process where an agent learns to make decisions by mapping specific situations to actions in order to maximize a cumulative reward. The agent is not given explicit instructions on which actions to take; instead, it must experiment with different actions to discover which ones generate the greatest rewards~\cite{sutton2018reinforcement}}. In many real-world domains like guided navigation and autonomous mission controls, the RL agent learns entirely from a dataset of past interactions rather than interacting in real-time with the environment~\cite{levine2020offline}. The dataset can be collected from agents following suboptimal or exploratory policies~\cite{meng2020qtaccel, levine2020offline}. The data logs include tasks such as \emph{Frozen Lake and Taxi}~\cite{brockman2016openai}. To illustrate, frozen lake environment involves crossing a frozen lake \ms{on foot} from start to goal without falling into \kailash{any holes}. The player may not always move in the intended direction due to the slippery nature of the frozen lake~\cite{brockman2016openai}. The taxi environment involves navigating to passengers in a grid world, picking them up, and dropping them off at one of four locations~\cite{brockman2016openai}. The end goal in this setting is still to optimize \kailash{a reward} function.

\begin{figure}[H]
\centering
\scalebox{1.0}{
\includegraphics[width=\linewidth]{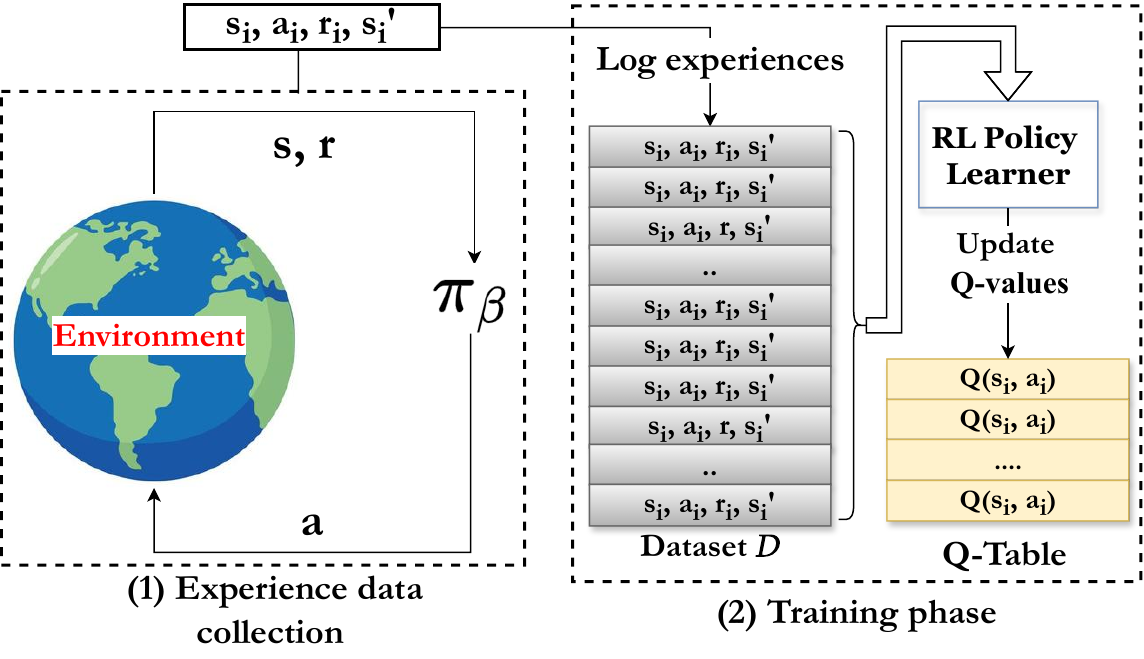}}
\caption{\kailash{The Operational Workflow of an Offline RL System. An unknown behavior policy \( \pi_\beta \) is used to gather the dataset, denoted as \( \mathcal{D} \). Training occurs without any interaction with the environment, and the policy~(i.e., final Q-table) is deployed only after it is fully trained. The architecture is adapted from~\cite{levine2020offline}.}}
\label{Figure1}
\end{figure}

To elaborate on the operational workflow of offline reinforcement learning, we illustrate the process in Figure~\ref{Figure1}.~We collect a large set of experiences using an unknown behavior policy\footnote[1]{\label{behavior_policy}A \emph{behavior policy} in reinforcement learning is the strategy an agent uses to explore and gather data from an environment~\cite{levine2020offline}} \( \pi_\beta \), and the obtained dataset is labeled as $\mathcal{D}$~\kailash{(Figure~\ref{Figure1}\circled{1})}. We perform this step once prior to the training phase of the reinforcement learning algorithm. In the offline training phase, the learning algorithm processes data tuples known as experiences\footnote[2]{We refer to experiences as transitions or samples interchangeably~\cite{levine2020offline}.} in the dataset, which includes states, actions, rewards, and next states~($s_i, a_i, r_i, s'_i$)~\cite{prudencio2023survey}. The \kailash{learning algorithm} uses this data to repeatedly update a policy, specifically refining the quality values associated with state-action pairs until the expected rewards are reached. This involves reading~(memory reads) from the dataset, learning, and then writing~(memory writes) these updated quality values associated to the state-action pairs to the Q-table~\kailash{(\circled{2})}.~After thoroughly training the policy~(i.e., constructing the final Q-table) for a number of episodes\footnote[3]{Note that each \emph{episode} involves computing the quality values and updating the \emph{quality} values associated with state-action pairs in the Q-table.}, the policy is then ready for testing and deployment.

\begin{figure}[H]
\centering
\scalebox{1.0}{
\includegraphics[width=\linewidth]{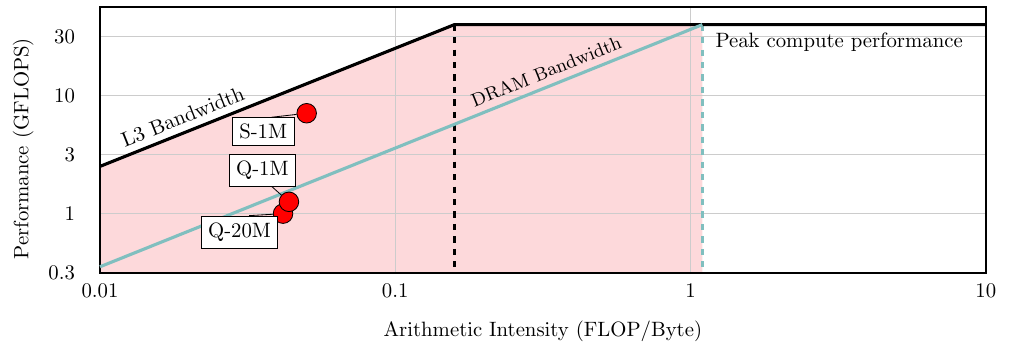}}
\caption{The Roofline model depicts the performance characteristics of CPU versions in RL workloads, where \textbf{``Q''} refers to \kailash{Q-learner~\cite{sutton2018reinforcement, watkins1992q, li2020sample}, \textbf{``S''} signifies \kailash{the} SARSA learner~\cite{sutton2018reinforcement}}, and ``1M'' and ``20M'' indicate the \kailash{data size in millions of transitions}. The workloads are tested on an Intel i7-9700K CPU.}
\label{Figure2}
\end{figure}

\kailash{RL algorithms typically exhibit memory-bounded behavior due to repeated memory accesses during the training phase. This involves iterating over the dataset multiple times to refine the policy and construct the final Q-table. To quantify the memory-boundedness of the CPU versions of our RL workloads~(Q-learning~\cite{sutton2018reinforcement, watkins1992q, li2020sample}, and SARSA learner~\cite{sutton2018reinforcement}), we employ a roofline model~\cite{williams2009roofline} to visualize the extent to which our workloads are constrained by memory bandwidth and computational limits. Figure~\ref{Figure2} shows the roofline model on an Intel(R) Core(TM) i7-9700K CPU~(Coffee lake) with Intel Advisor~\cite{Intel_advisor}.}

\kailash{The shaded area at the intersection of DRAM bandwidth and the peak compute performance roof is defined as the memory-bound area in the roofline plot. We make a key observation from Figure~\ref{Figure2} that both the Q-learner and SARSA-learner CPU versions are in the memory-bound region. This is because their performance is primarily constrained by low DRAM bandwidth, which prevents the RL algorithms from achieving the maximum possible hardware performance. As a result, these RL workloads are potentially suitable for PIM.}


\subsection{Processing-In-Memory}
\label{PIM}

\begin{figure}[H]
\centering
\scalebox{1.0}{
\includegraphics[width=\linewidth]{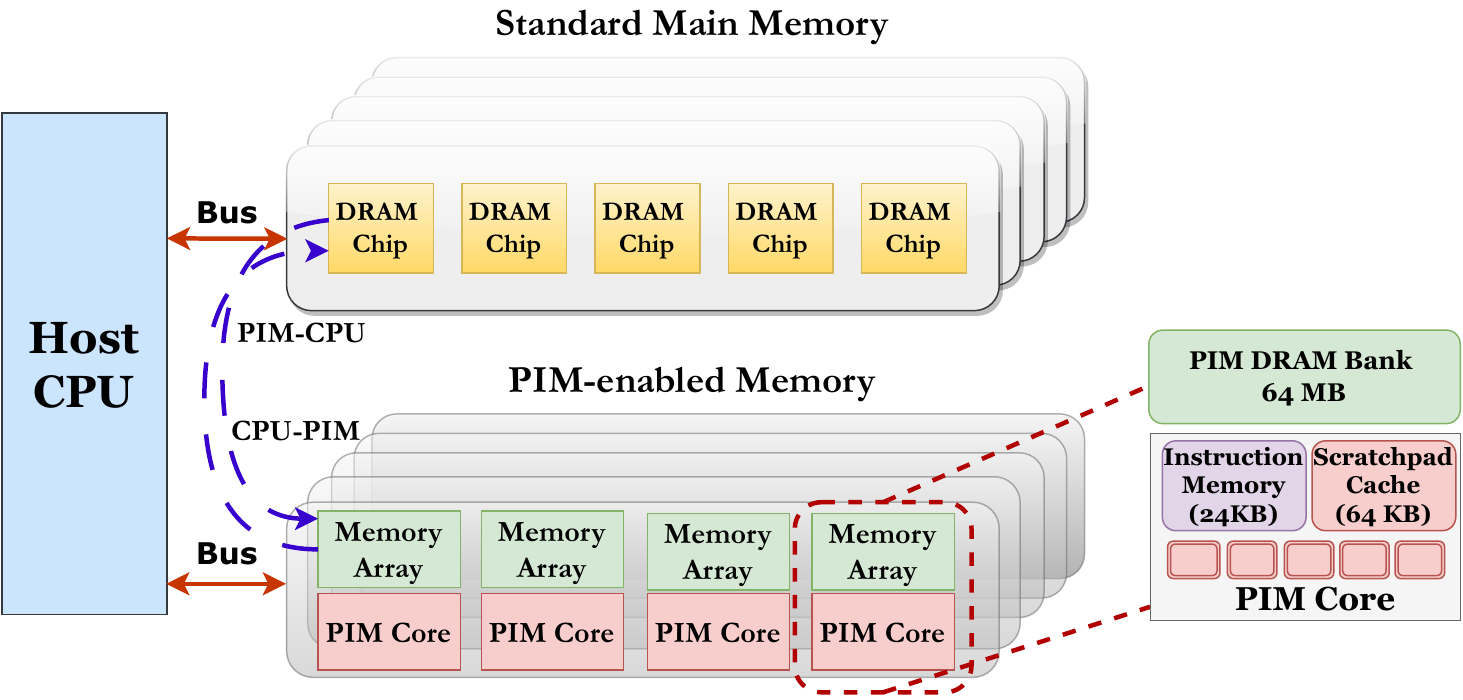}}
\vspace{-\baselineskip}
\caption{\kailash{Organization of a State-of-the-Art Processing-In-Memory (PIM) Architecture.} \kailash{Adapted from~\cite{gomez2021benchmarking, gomez2023evaluating, chen2023simplepim, UPMEM, giannoula2022towards}.}}
\label{Figure3}
\end{figure}

\kailash{Recently, real-world Processing-In-Memory (PIM)~\cite{gomez2021benchmarking, chen2023simplepim, gomez2022benchmarking, giannoula2022towards, oliveira2023transpimlib, nider2021case, gomez2023evaluating, das2022implementation, liu2018processing, mutlu2022modern, abecassis2023gapim, diab2022high,rhyner2024analysis, giannoula2024accelerating, giannoula2022sparsep, diab2023framework, gupta2023evaluating, fernandez2024matsa, fernandez2022accelerating, gomez2022machine, hyun2024pathfinding, bernhardt2023pimdb, noh2024pid, chen2023uppipe,khan2022cinm,zois2018massively,kang2023pim,falevoz2023energy,lopes2024pim,mognolparallelization,kim2023virtual, nider2022bulk, nider2020processing} systems have emerged and are now part of the market landscape, with UPMEM~\cite{gomez2021benchmarking, chen2023simplepim, gomez2022benchmarking, giannoula2022towards, oliveira2023transpimlib, nider2021case, gomez2023evaluating, das2022implementation, liu2018processing, mutlu2022modern, abecassis2023gapim, diab2022high,rhyner2024analysis, giannoula2024accelerating, giannoula2022sparsep, diab2023framework, gupta2023evaluating, fernandez2024matsa, fernandez2022accelerating, gomez2022machine, hyun2024pathfinding, bernhardt2023pimdb, noh2024pid, chen2023uppipe,khan2022cinm,zois2018massively,kang2023pim,mognolparallelization,kim2023virtual,lopes2024pim,falevoz2023energy,baumstark2023adaptive, ferraz2023unlocking,lavenier2016blast,ma2024accelerating,wu2023pim,khan2024landscape} pioneering the first-ever commercialization of a PIM architecture.} Additionally, there have been announcements regarding Samsung HBM-PIM~\cite{kwon202125, lee2021hardware}, Samsung AxDIMM~\cite{ke2021near}, SK Hynix AiM~\cite{lee20221ynm}, and Alibaba HB-PNM~\cite{niu2022184qps}. All these architectures have been prototyped and evaluated on real systems, sharing key and significant characteristics, as illustrated in~Figure~\ref{Figure3} \ms{for the UPMEM PIM system.} First, these PIM systems feature a host CPU processor integrated with standard main memory, a deep cache hierarchy, and PIM-enabled memory modules. Second, the PIM-enabled memory contains multiple PIM chips connected to the host CPU through a memory channel. Third, the PIM processing elements operate at relatively low frequencies, typically a few hundred MegaHertz. Each PIM core (\kailash{i.e., processing} element\ms{;} PIM PE) may include a small private instruction memory and a small data storage (scratchpad cache) memory. Fourth, PIM PEs can access data in their local memory bank, and there is typically no direct communication channel among the PIM cores. However, communication between multiple PIM cores occurs typically through the host CPU processor in architectures like UPMEM~\cite{UPMEM, UPMEM_1,gomez2022benchmarking}, HBM-PIM~\cite{kwon202125, lee2021hardware}, and AiM~\cite{lee20221ynm}. 

In this study, we use UPMEM PIM, which uses conventional 2D DRAM arrays and tightly integrates them with general-purpose PIM cores, namely DRAM Processing Units~(DPUs), on the same chip. 
UPMEM-based PIM systems use the Single Program Multiple Data
(SPMD)~\cite{mutlu2020lecture} programming model. The DPUs are programmed using the C language with additional library calls~\cite{UPMEM, UPMEM_1, UPMEM_2, UPMEM_3}, while the host library offers C++, Python, and Java API support. The UPMEM SDK~\cite{UPMEM_3} supports common C data types and interfaces seamlessly with the LLVM compilation framework~\cite{lattner2004llvm}. For a comprehensive listing of supported instructions, we refer the reader to the UPMEM PIM user manual~\cite{UPMEM_2}.

The state-of-the-art UPMEM architecture has 20 PIM-enabled DIMMs, each with two ranks of 8 PIM chips. In total, 2,560 DPUs~(PIM cores) are deeply pipelined and implemented with fine-grained multi-threading~\cite{smith1986pipelined, smith1982architecture} providing a peak throughput of 1 TOPS~($Tera$ $operations/second$). Each PIM chip has eight 64-MB DRAM banks, with a programmable PIM core, 24-KB instruction memory~(IRAM), and a 64-KB scratchpad memory~(WRAM) coupled to each bank~\cite{giannoula2022towards}. The DPUs have in-order 32-bit RISC-style instruction set architecture operating at 450 MHz~\cite{UPMEM_1, UPMEM}. The DPU has 24 hardware threads, each with 24 32-bit general-purpose registers. They include native support for 32-bit integer addition/subtraction and 8-bit multiplication. The complex operations, such as the multiplications on 64-bit integers, 32-bit 
floating point operations, and 64-bit 
floating point operations, are emulated by the run-time library 
and take tens to thousands of cycles~\cite{giannoula2022towards, gomez2023evaluating, gomez2021benchmarking, gomez2022benchmarking}. 

The conventional main memory and PIM-enabled memory modules exhibit distinct data layouts. The host processor can access MRAM banks for tasks such as copying input data (from main memory to MRAM, i.e.,~CPU to DPU) and retrieving results (from MRAM to main memory, i.e., DPU to CPU). Since there are no direct communication channels between DPUs, inter-DPU communication occurs exclusively through the host CPU, utilizing parallel CPU-DPU and DPU-CPU data transfers~\cite{gomez2023evaluating, giannoula2022towards,lee2024analysis}.
 
Even though we demonstrate adaptation of RL workloads to UPMEM architecture, our proposed optimization strategies are versatile and can be deployed on other real PIM hardware, resembling the architecture illustrated in Figure~\ref {Figure3}. Thus, we use the terms PIM core, PIM thread, DRAM bank, scratchpad, and CPU-PIM/PIM-CPU data transfer, which correspond to the DPU, tasklet, MRAM bank, WRAM, and CPU-DPU/DPU-CPU transfer in the PIMs implemented by UPMEM. \kailash{For more detailed analysis of the UPMEM architecture, we refer the reader to~\cite{hyun2024pathfinding,gomez2022benchmarking,gomez2021benchmarking,giannoula2022towards}.}

\section{\name Design and Implementation}
\label{Implementation}

In this section, we \kailash{first} study the memory behavior of RL workloads and their performance bottlenecks. \kailash{We then} demonstrate our workload adaptation strategies for PIM. 

\subsection{Memory Behavior of RL workloads}
\label{memory_profile}
The training phase of offline reinforcement learning is heavily influenced by the need to access experiences stored in a dataset, which were gathered using a specific behavior policy. This phase often encounters memory bottlenecks due to two primary reasons:~\circled{1}~RL algorithms sequentially process large volumes of historical experience data for optimal policy learning, and~\circled{2} Different sampling strategies~(impacting data locality) are used during the learning process. For instance, complex environments~(e.g., Atari~\cite{mnih2013playing}, StarCraft~\cite{vinyals2017starcraft, mei2023mac}) typically require the agent to explore a broad range of \kailash{the} state-action space in early time-steps, so the agent performs random sampling. This random sampling can result in irregular memory access patterns and poor data locality as the state-action space expands~\cite{gogineni2023accmer, gogineni2023towards, gogineni2023scalability}. To mitigate these bottlenecks, we first implement our workloads with different sampling strategies and test the efficiency of PIM. We distribute data chunks across various PIM cores in memory to accelerate the training phase and execute batch updates for each iteration \kailash{in} near-bank PIM \kailash{cores}~\cite{gomez2021benchmarking, gomez2023evaluating, gomez2022benchmarking}.



\subsection{Implementation of RL algorithms on PIM Architecture}
\label{Impleemntation_pim}
\kailash{Tabular Q-learning~\cite{sutton2018reinforcement, watkins1992q, li2020sample} and SARSA~\cite{sutton2018reinforcement} are popular RL algorithms widely used in various applications~\cite{perera2021applications,jang2019q,peng1994incremental,sharma2020deep,spano2019efficient,applebaum2022bridging,perkins2002existence,rummery1994line,shresthamali2017adaptive} and as part of machine learning for hardware/software systems~\cite{bera2021pythia,ipek2008self,pd2015q,wang2019high,singh2022sibyl}. Both the algorithms learn from Q-tables. The Q-tables store the \textit{quality} values associated with state-action pairs.}

\subsubsection{\textbf{Q-learning}}
\label{Tabular_Q}
Tabular Q-learning~\cite{sutton2018reinforcement, watkins1992q, li2020sample} is a widely-used model-free and \kailash{off-policy RL workload~\cite{sutton2018reinforcement, watkins1992q, li2020sample}} that learns through a trial-and-error approach\kailash{~\cite{sutton2018reinforcement, watkins1992q, li2020sample,perera2021applications,jang2019q,peng1994incremental,sharma2020deep,spano2019efficient,applebaum2022bridging, perkins2002existence}}. \kailash{Agents} interact with the environment based on some \kailash{off-policy approach~\cite{sutton2018reinforcement, levine2020offline}} like \emph{random selection} or \emph{epsilon greedy}. In order to train the workload offline, we employ \kailash{a behavior policy~(i.e., random action selection)~\cite{sutton2018reinforcement, levine2020offline}} to collect the dataset $\mathcal{D}$ \kailash{once}. While we use the random action selection, other policies such as \emph{epsilon greedy} and \emph{boltzmann} can also be used to execute actions on the environment and log the experiences~\cite{langford2007epoch, sutton2018reinforcement, caspi_itai_2017_1134899}. The objective in this offline setting is still to collect enough experiences and learn a policy~(i.e., constructing the final Q-table) that maximizes the expected return~\cite{sutton2018reinforcement}.

Each experience tuple in the dataset is represented as $\mathcal{D}_i$ = {($s_i, a_i, r_i, s'_i$)}, where $i$ denotes the index of the transition within the dataset. $s'$ represents the next state resulting from taking action $a$ in state $s$, and the reward $r$ is determined by the state-action pair $(s,a)$.~\kailash{$\gamma$ \kailash{is a} discount factor that determines the balance between the immediate and future rewards~\cite{sutton2018reinforcement}}. \kailash{$\alpha$ is the learning rate parameter that determines the rate at which the quality values associated with the state-action pairs in the Q-table are updated~\cite{sutton2018reinforcement}.} 

To illustrate the offline training phase of the tabular Q-learning algorithm, we outline the steps in Algorithm~\ref{Algorithm1}.~The Q-learning algorithm initializes a Q-table with arbitrary values. For each episode~(line 5), multiple batches~(line 7) are selected to iteratively update the Q-values~(line 12) based on the total number of experiences in the dataset~\cite{sutton2018reinforcement}.~\kailash{The term $\max_{a'} Q(s', a')$ is the maximum Q-value for the next state $s'$, across all possible actions $a'$~(line 10). In other words, it calculates the highest Q-value  for the next state $s'$, by trying out all the available actions $a'$, ensuring the selection of the most rewarding action for the next state.}

\kailash{We} note that multi-agent reinforcement learning has \ms{recently} been widely adopted in several popular domains from gaming to autonomous \kailash{driving~\cite{de2020independent, zhou2021smarts,shalev2016safe, palanisamy2020multi, bhalla2019training,ozdaglar2021independent,kiran2021deep,yang2018cm3,matignon2012independent,gu2022review}}. In this decentralized Q-learning approach, each agent maintains its own experience dataset and updates its Q-table to make decisions~\cite{tan1993multi}. 
In our modeling, the independent learners do not observe the actions of other agents in the system. 

\begin{algorithm}[ht]
\caption{Tabular Q-learning Algorithm}
\label{Algorithm1}
\begin{algorithmic}[1]
\State \textbf{Inputs:}
\State Experience data collected offline.
\State Initialize \kailash{a} Q-table with arbitrary/zero values.
\State Hyper-parameters: Learning Rate ($\alpha$), Discount Factor ($\gamma$), $num\_episodes$.
\For {each $episode$ in $num\_episodes$}
    \State \textbf{Batched updates:}
    \For {each \kailash{experience~($s_i, a_i, r_i, s'_i$)} in selected batch}
            \State \textbf{Q-learning Update:}
            \State Calculate Q-value target for the experience:
            \kailash{\State $q_{\text{value target}} \leftarrow \text{r} + \gamma \cdot \max_{a'} Q(s', a')$
            \State Update Q-values for the current state-action:
            \State $Q(s, a) \leftarrow Q(s, a) + \alpha \cdot (q_{\text{value target}} - Q(s, a))$}
    \EndFor
\EndFor
\State \textbf{Output:}
\State Final Q-table with the learned Q-values.
\end{algorithmic}
\end{algorithm}

\begin{figure*}[t]
\centering
\scalebox{1.0}{ 
\includegraphics[width=\linewidth]{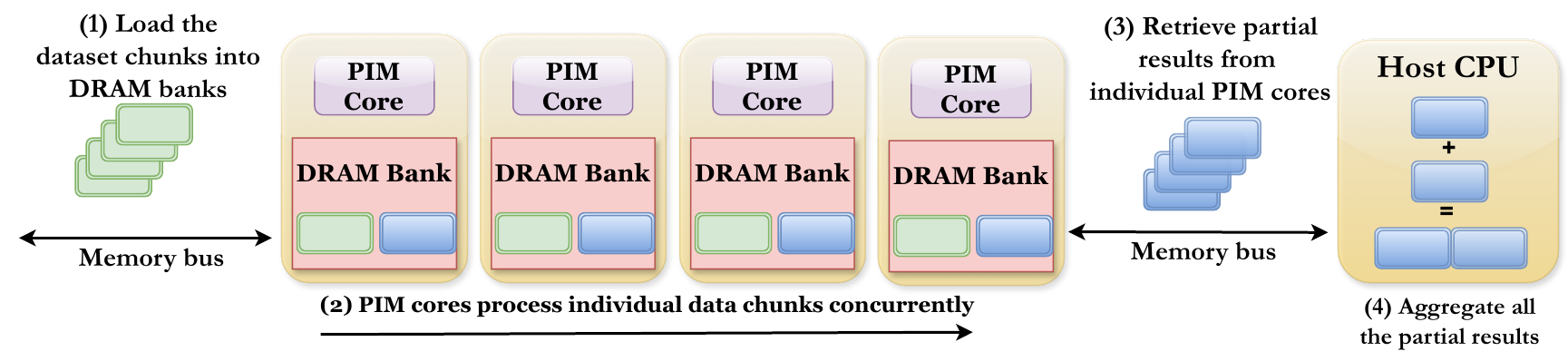}}
\caption{\kailash{\name execution on a real PIM system. The execution phase comprises four main steps: (1) loading the input dataset chunks into individual DRAM banks of PIM-enabled memory, (2) executing the RL workload (kernel) on PIM cores in parallel operating on different chunks of data, (3) retrieving partial results from DRAM banks to the host CPU, and (4) aggregating partial results on the host processor.}}
  \label{Figure4}
\end{figure*}

\textbf{PIM Implementation:}~
Initially, the training dataset ($\mathcal{D}$) resides in the main memory of the host CPU. The first step involves transferring individual chunks of the training dataset to the local memories (DRAM banks) of PIM cores. To maximize parallelism in the Q-value update kernel, we partition the training dataset ($\mathcal{D}$) so that each PIM core ($\mathcal{P}_i$) handles a distinct chunk of data ($\mathcal{C}_D$), enabling faster memory accesses. Secondly, within each PIM core ($\mathcal{P}_i$), Q-learning updates~(line 12 from Algorithm~\ref{Algorithm1}) are computed for each transition ($\mathcal{T}_i$) in ($\mathcal{C}_D$) using a single hardware thread~(this work focuses solely on PIM-core parallelism). Each PIM core ($\mathcal{P}_i$) is allocated a Q-table with arbitrary values, and all PIM cores train in parallel asynchronously updating their local Q-tables using the \textit{state-action-reward-next state} trajectories. The third step involves transferring partial results obtained from processing the individual data chunk in a specific PIM core back to the host processor to aggregate the final Q-table. \kailash{The operational workflow of \name execution on a real PIM system is described in~Figure~\ref{Figure4}}.

We implement six versions of Q-learning with different input data types, and  three sampling strategies (\emph{SEQ} - sequentially sampling experiences, \emph{RAN} - randomly selecting experiences to update for thorough exploration, and \emph{STR} - selecting experiences at regular intervals). We note that experimenting with different sampling strategies enables us to assess their impact on the computational workload during the sampling phase.

\begin{itemize}
    \setlength\itemsep{0.2em}
    \item {\fontfamily{lmtt}\selectfont
Q-learner-FP32
} trains with 32-bit real values for  $\max_{a'} Q(s', a')$, and Q${\text{-table}}$ initialization and no scaling optimization. We use 32-bit data types as they provide a more accurate representation of transition data across episodes than 16-bit data.

    \item {\fontfamily{lmtt}\selectfont
Q-learner-INT32
} trains with 32-bit fixed point representations of $\max_{a'} Q(s', a')$, Q${\text{-table}}$ initialization, and we scale up the reward $r$ for each experience~$\mathcal{T}_i$, learning rate $\alpha=0.1$, and the discount factor $\gamma=0.95$ by using a constant scale factor=10,000~(scale factor is chosen to prevent overflow and underflow errors while ensuring sufficient precision for floating-point multiplications) to mitigate the cost of floating-point multiplications for the update equation~(line 12) in Algorithm~\ref{Algorithm1}. We scale down after the Q-value update and finally store the descaled value in the Q-table. This hybrid implementation is motivated by the fact that real-world PIM cores only support arithmetic operations of limited precision. For instance, UPMEM DPUs~\cite{UPMEM_2} execute naive 8-bit integer multiplication and emulate the 32-bit integer multiplication using shift-and-add instructions~\cite{gomez2022benchmarking, gomez2021benchmarking}. Apart from UPMEM, other accelerators like HBM-PIM~\cite{kwon202125} and AiM~\cite{lee20221ynm} feature only 16-bit floating point operations. Additionally, replacing the compiler-generated 16-bit and 32-bit multiplications with \textit{custom 8-bit built-in multiplications}~\cite{gomez2023evaluating} may be adopted to boost the training time further and reduce the number of instructions, but this optimization, which is specific to UPMEM, might only apply to some environments~(e.g., frozen lake) which have limited value range that fits in 8 bits. Additionally, we implement custom routines such as \emph{linear congruential generator~\cite{10.1145/318123.318229}} to replicate the functionality of the \texttt{rand()} function within PIM cores~(some standard library functions are not supported by UPMEM PIM architecture).
    \item We evaluate the performance of the aforementioned two versions using different sampling strategies with the data laid out in diverse memory access patterns. The six implemented versions include:~{\fontfamily{lmtt}\selectfont Q-learner-SEQ-FP32, Q-learner-RAN-FP32, Q-learner-STR-FP32, Q-learner-SEQ-INT32, Q-learner-RAN-INT32, Q-learner-STR-INT32}.
\end{itemize}

\paragraph{\textbf{Multi-agent Q-learning}}~For multi-agent Q-learning, we employ a random policy to explore the environment and log individualized experiences. Subsequently, we load the agent-specific datasets into the PIM cores' local memories (DRAM banks). The only difference in this workload compared to the Q-learning is that each PIM core will have agent-specific experiences to learn from, enabling multiple independent learners concurrently. We pin each agent to a PIM core in this design and iteratively train on its unique dataset. The host processor retrieves the final Q-tables for multiple agents upon completion of the training process. Notably, the aggregation step would be unnecessary in this setting as the learners operate independently throughout the training.

\subsubsection{\textbf{SARSA Learning}}
\label{SARSA}
SARSA~(State-Action-Reward-State-Action) is an on-policy algorithm that learns the optimal policy by continuously updating the policy toward achieving the maximum reward~\cite{sutton2018reinforcement}.~\kailash{The only difference for SARSA compared to Q-learning is that, SARSA employs an epsilon-greedy approach~\cite{sutton2018reinforcement} to select next action $a'$~(Equation~\ref{eq1}) and this function uses a custom routine~(\texttt{rand()} function) to generate random action~\cite{sutton2018reinforcement}.} \newline
The SARSA update equation is:
\begin{equation}
\label{eq1}
    \kailash{Q(s, a) \leftarrow Q(s, a) + \alpha \left( \text{r} + \gamma Q(s', a') - Q(s, a) \right)}
\end{equation}
The terms are similar to Q-learning, and the SARSA algorithm has the same training pattern, but the key difference is $Q(s', a')$ term, where the action $a'$ is the actual next action taken, following the policy being learned. Instead, in Q-learning maximum Q-value across all possible actions is used~\cite{sutton2018reinforcement}.



\textbf{PIM Implementation:}~To extract parallelism in the SARSA learning update kernel, the first step involves partitioning the training dataset ($\mathcal{D}$) into subsets of equal size so that each PIM core ($\mathcal{P}_i$) receives a unique chunk of data ($\mathcal{C}_D$). Secondly, a hardware thread is dedicated to computing SARSA updates~(Equation~\ref{eq1}) within each PIM core ($\mathcal{P}_i$). We note that, similar to the Tabular Q-learning implementation, we currently focus on the core-level parallelism in this work. In the third step, the results obtained from processing individual chunks of data within each PIM core ($\mathcal{P}_i$) are then transferred back to the host processor. The host processor finally aggregates all the partial results from multiple PIM cores, facilitating the final Q-table learned using SARSA update rule~\cite{sutton2018reinforcement}. SARSA learner follows the same arithmetic intensity as Q-learning since only one floating-point multiplication is needed, and we substitute it with the fixed-point representation and scaling optimization. 

We implement different variations of SARSA learning, featuring different data types~(FP32 and INT32) and diverse sampling strategies. {\fontfamily{lmtt}\selectfont
SARSA-learner-INT32} learning uses 32-bit fixed point representations of $Q(s', a')$, $Q{\text{-table}}$ initialization, and we scale up the reward $r$ for each experience~$\mathcal{T}_i$, learning rate $\alpha=0.1$, and the discount factor $\gamma=0.95$ by using a constant scale factor=10,000 and scale down after the experience update and finally transfer Q-values back to the host CPU in original precision. We implement the {\fontfamily{lmtt}\selectfont SARSA-learner-INT32} as a substitute for the naive version~({\fontfamily{lmtt}\selectfont SARSA-learner-FP32}) that trains with 32-bit real values~(for variables mentioned above), which takes relatively longer execution cycles as the UPMEM PIM only supports native integer multiplications~\cite{UPMEM, gomez2021benchmarking, gomez2022benchmarking, gomez2023evaluating}~(specifically 8-bit). We evaluate the performance of the two variations mentioned above using diverse sampling strategies~(\emph{SEQ} - sequentially sampling experiences, \emph{RAN} - randomly selecting experiences to update for thorough exploration, and \emph{STR} - selecting experiences at regular intervals). The six implemented versions for SARSA are:~{\fontfamily{lmtt}\selectfont SARSA-SEQ-FP32, SARSA-RAN-FP32, SARSA-STR-FP32, SARSA-SEQ-INT32, SARSA-RAN-INT32,~SARSA-STR-INT32}.

\section{Experimental Evaluation}

\kailash{In this section, we first describe our experimental setup. Second, we evaluate \name in terms of training quality~(Section~\ref{Quality}), and performance scaling characteristics, specifically strong scaling results~(Section~\ref{Performance_analysis_PIM_kernel}). Third, we compare our PIM implementations to state-of-the-art CPU and GPU implementations~(Section~\ref{comparison_cpu_gpu}).}

\label{Evaluation}
\subsection{Experimental Setup}
\label{setup}
Table~\ref{tab:system-specs} summarizes the specifications of the UPMEM PIM, CPU, and GPU systems. We \kailash{perform our} experiments on a real-world PIM server [156] with 2,524 PIM cores running at 425 MHz and 158 GB of DRAM memory. The table also outlines the characteristics of the baseline CPU and GPU systems used for comparative analysis.
\begin{figure*}[t]
\centering
\scalebox{1.0}{
\includegraphics[width=17cm, height=15cm]{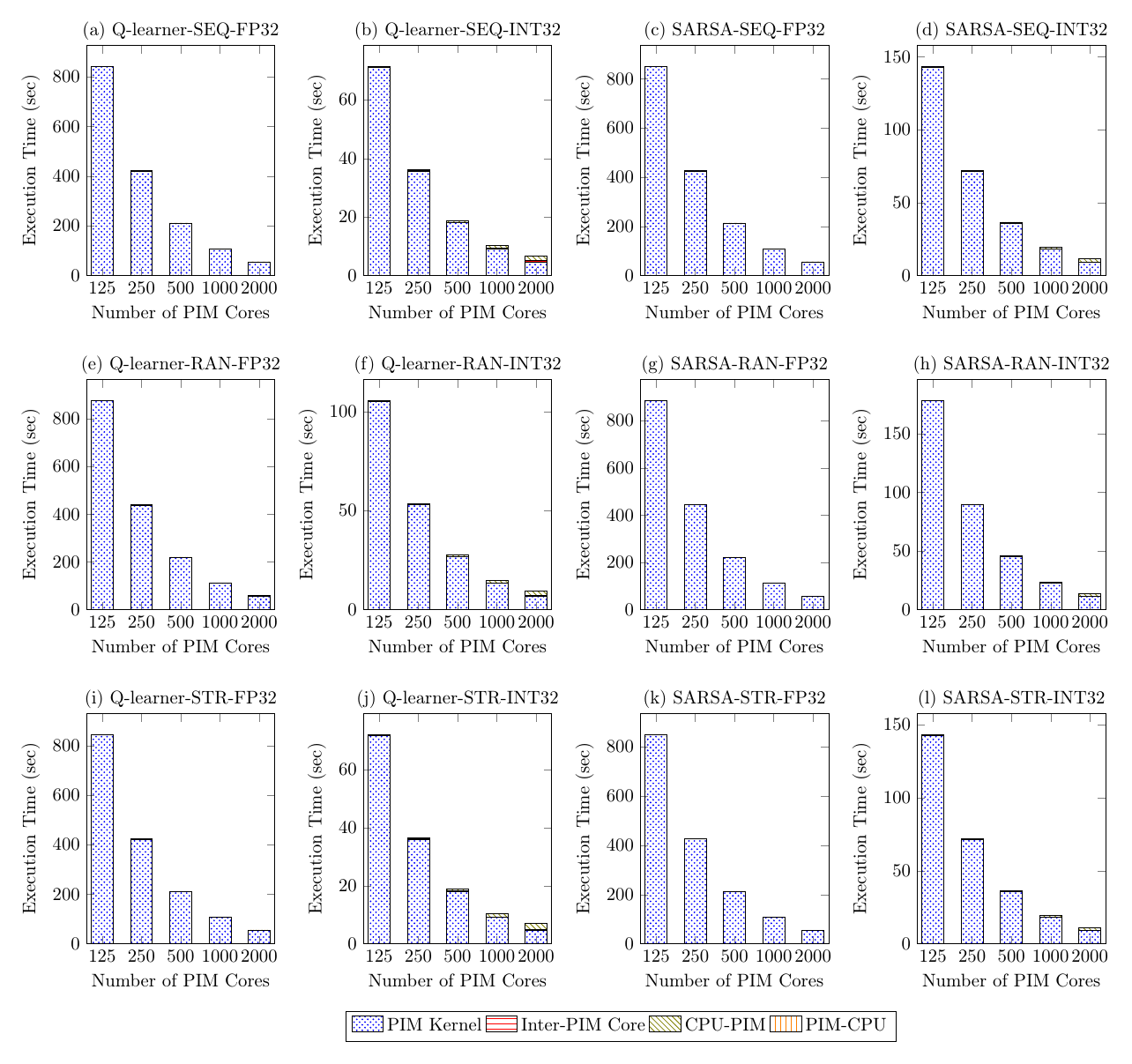}}
\caption{Execution time (measured in seconds) of RL workloads on 125, 250, 500, 1,000  and 2,000 PIM cores~(x-axis) with each PIM core running with single thread for a frozen lake environment. In this illustration, SEQ-RAN-STR represent sequential, random and stride-based sampling techniques and both the Q-learning and SARSA learners are evaluated with FP32 and INT32 data types. The synchronization period $\tau$ and the stride value in this experiment is set to 50 and 4 respectively.}
\label{Figure5}
\end{figure*}

\begin{table}[H]
\centering
\caption{Evaluated UPMEM PIM system~\cite{gomez2021benchmarking, gomez2022benchmarking, gomez2023evaluating} and baseline CPU~\cite{CPU} and GPU~\cite{GPU} specifications.}
\begin{tabular}{ |>{\centering\arraybackslash}m{5em}||>{\centering\arraybackslash}m{1.5cm}|>{\centering\arraybackslash}m{2cm}|>{\centering\arraybackslash}m{2cm}| } 
  \hline
  \textbf{Metric} & \textbf{UPMEM PIM System} & \textbf{Intel Xeon Silver 4110 CPU~\cite{CPU}} & \textbf{NVIDIA Ampere RTX 3090 GPU~\cite{GPU}} \\ 
  \hline
  \hline
  Processor Node & 2x nm & 14 nm & 8 nm \\
  \hline
  Total Cores & 2,524 & 8 (16 threads) & 82 cores \newline \scriptsize (10496 SIMD lanes) \\
  \hline
  Frequency & 425 MHz & 2.4 GHz \newline \scriptsize (3 GHz Turboboost) & 1.70 GHz \\
  \hline
  Peak \newline Performance & 1,088 GOPS & 38 GFLOPS & 35,580 GFLOPS \\
  \hline
  Main Memory & 158 GB & 132 GB & 24 GB \\
  \hline
  Memory Bandwidth & 2145 GB/s & 28.8 GB/s & 	936.2 GB/s \\
  \hline
  Component TDP & 280 W & 85 W  & 	350 W \\
  \hline 
\end{tabular}
\label{tab:system-specs}
\end{table}

The RL workloads are evaluated using popular environments, namely frozen lake and taxi, developed by Gym~\cite{brockman2016openai}. The taxi environment has a state space of $Discrete(500)$ since there are 25 taxi positions, 5 possible locations of the passenger (including the case when the passenger is in the taxi), 4 destination locations, and an action space of $Discrete(6)$, while the frozen lake environment has a state space of $Discrete(16)$ since the map size is $4\times4$ and an action space of $Discrete(4)$. In our experiments, we use a learning rate of 0.1, $\gamma$-the discount factor, set to 0.95, and train the workloads for 2,000 episodes. To obtain a partially trained policy, we train a random behavior policy online and log the experiences until the policy performance achieves a performance threshold (Average reward) for frozen lake and taxi, respectively. We collected 1 million transitions for frozen lake and 5 million for the taxi paradigm. We collected more data for the taxi environment~\cite{sutton2018reinforcement} because it encompasses $31.25\times$ more states compared to the frozen lake envrionment~\cite{sutton2018reinforcement, brockman2016openai}.

\begin{figure*}[t]
\centering
\scalebox{1.0}{
\includegraphics[width=17cm, height=15cm]{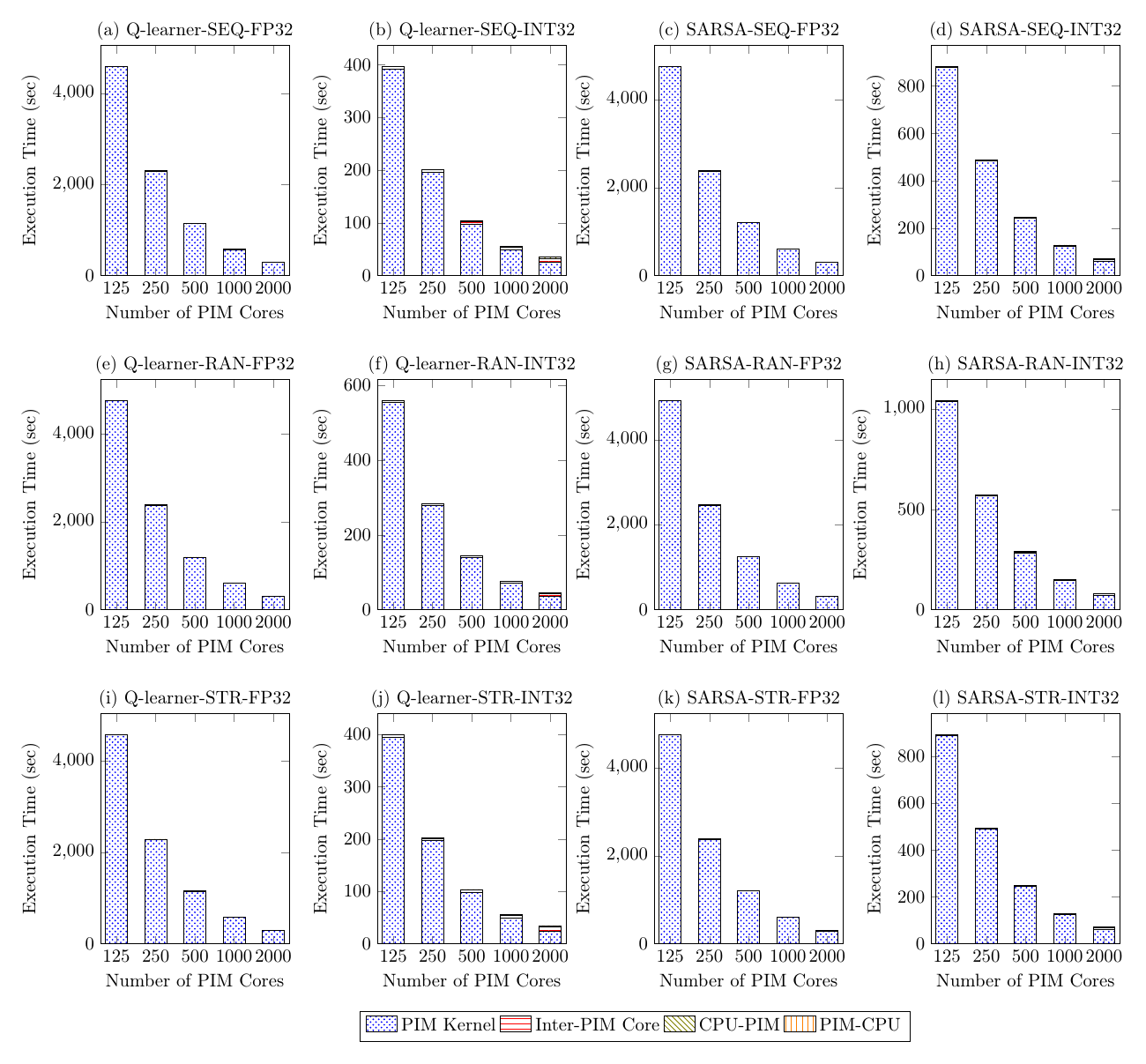}}
\caption{Execution time (sec)~(y-axis) of RL workloads on 125, 250, 500, 1,000 and 2,000 PIM cores~(x-axis) with each PIM core running with single thread for a taxi environment. In this illustration, SEQ-RAN-STR represent sequential, random and stride-based sampling techniques and both the Q-learning and SARSA learners are evaluated with FP32 and INT32 data types. The synchronization period $\tau$ and the stride value in this experiment is set to 50 and 4 respectively.}
\label{Figure6}
\end{figure*}


\subsection{RL Training Quality}
\label{Quality}

The trained policy in Q-learning and SARSA is evaluated with the frozen lake and taxi environments. The hyper-parameters for the evaluation include 1,000 episodes with synchronization period ($\tau$) set to 50~\cite{woo2023blessing,qi2021federated}. $\tau$ refers to the communication rounds for the Inter-PIM core communication, where the total number of episodes, denoted by $\mathcal{E}$, is assumed to be divisible by $\tau$. The algorithm outputs the final aggregated Q-estimate as the average of all local Q-tables. In this context, $Comm_{rounds}$ is defined as $\mathcal{E} / \tau$, representing the rounds of communication required in the training phase. Our performance results of PIM implementations takes into account the estimated time, which includes the impact of $Comm_{rounds}$, specifically in the context of Inter-PIM core communication.

For the frozen lake environment, for {\fontfamily{lmtt}\selectfont Q-learner-SEQ}, we estimate the average mean reward for 1000 episodes with synchronization period~($\tau$) set at 10, 25, and 50 is observed to be 0.74, 0.7295, and 0.70, respectively. These are relatively same or slightly better than CPU implementation. 
For the {\fontfamily{lmtt}\selectfont SARSA-learner-SEQ} with a $\tau$ of 50, a mean reward of 0.71 is registered against the 0.723 of the CPU version. We note that the {\fontfamily{lmtt}\selectfont Q/SARSA-learner-RAN/STR} also perform on par with {\fontfamily{lmtt}\selectfont Q/SARSA-learner-SEQ}. 

For the taxi environment, we evaluate an approximated model {\fontfamily{lmtt}\selectfont Q-learner-SEQ} algorithm with synchronization period~($\tau$) set at 50 is observed to be -7.9 against the -8.6 of the CPU version. The {\fontfamily{lmtt}\selectfont SARSA-learner-SEQ} exhibits similar behavior with $\tau$ of 50, a mean reward of $-8.8$ against the $-8.2$ of the CPU version. 
Even with INT32, we convert the values back from INT32 to FP32 using scaling optimization before the PIM cores transfer the partial results to the host CPU.

\begin{figure*}[ht]
\centering
\scalebox{1.0}{
\includegraphics[width=\linewidth, height=8cm]{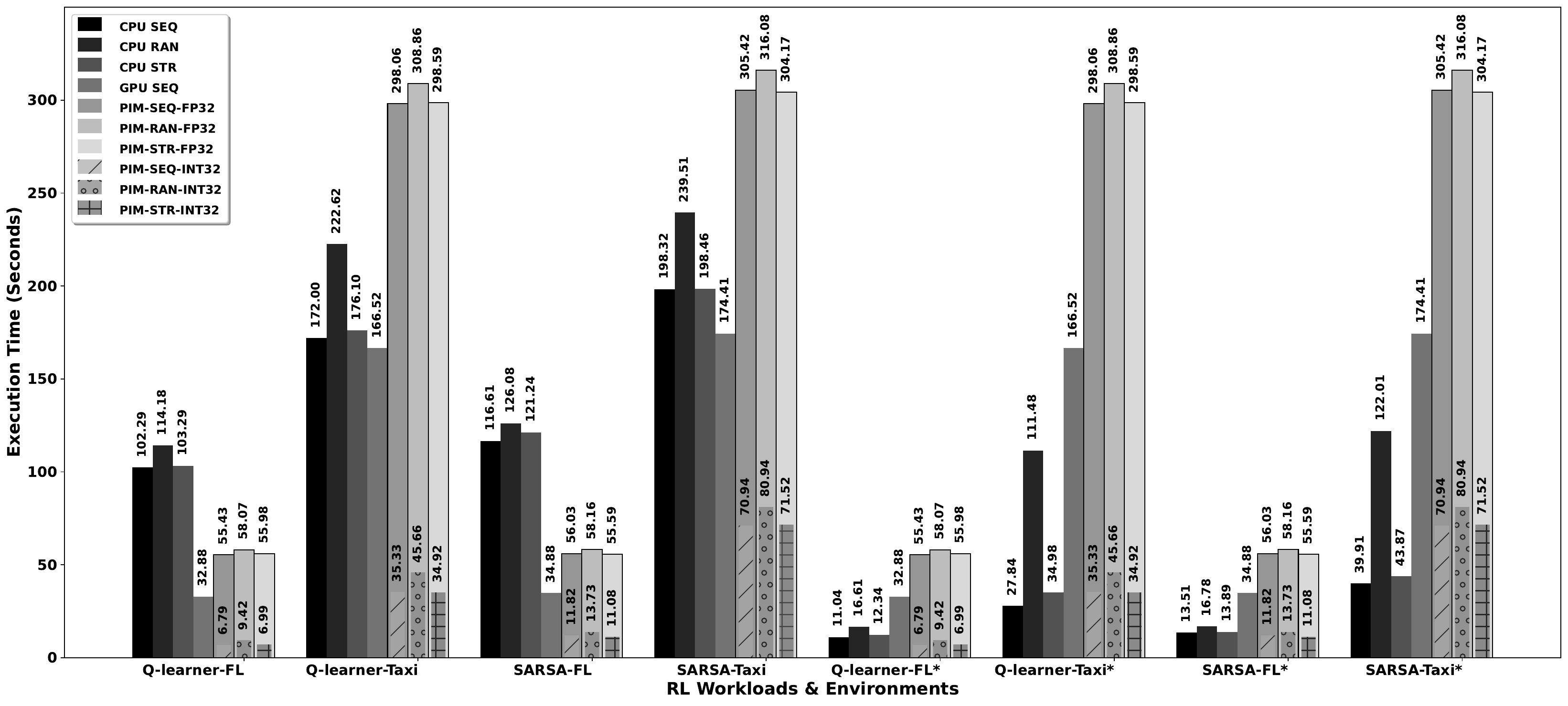}}
\caption{Execution time (in seconds) of CPU, GPU, and PIM for the training phase of RL workloads. The solid bars represent PIM execution time without scaling optimization, utilizing FP32. Meanwhile, the patterned bars depict execution time with scaling optimization, employing INT32. For PIM evaluation, we utilize 2,000 PIM cores~(best-performing number) for workload execution. The results for performance comparison with CPU-V2 version, Q-learning, and SARSA with FL and taxi are denoted in (*). The RL algorithms, {\fontfamily{lmtt}\selectfont Q-learner-FL, Q-learner-Taxi, SARSA-FL, and SARSA-Taxi}, refer to Tabular Q-learning and SARSA algorithms trained with frozen lake and taxi environments.}
\label{Figure7}
\end{figure*}


\subsection{Performance Analysis of PIM Kernels: Scaling PIM Cores}
\label{Performance_analysis_PIM_kernel}

In this section, we evaluate the performance scaling characteristics of our RL workloads using {\it strong scaling} experiments. 
Figure~\ref{Figure5} illustrates the performance scaling results across 125-2000 PIM cores for various versions of our RL workloads. We present the total execution time, which is further broken down into (1) the execution time of the PIM kernel, (2) communication time between the host processor and PIM cores for initial dataset transfer (CPU-PIM), (3) communication time between PIM cores and the host CPU for final result transfer~(PIM-CPU), and (4) communication time between PIM cores for Q-values transfer~(Inter PIM core). The Inter PIM core synchronization is estimated using the synchronization period ($\tau$), where the total number of episodes, denoted by $\mathcal{E}$, is assumed to be divisible by $\tau$. Given that we train for 2,000 episodes with $\tau$ set to 50, the value of $Comm_{rounds}$ is 40. During this process, the local results are aggregated before advancing to next episode. 

\kailash{To demonstrate how the performance of RL workloads scales with an increasing number of PIM cores, we maintain a fixed dataset size for the frozen lake and taxi environments~\cite{brockman2016openai}.}
\kailash{We make four observations from Figure~\ref{Figure5} and Figure~\ref{Figure6}. First,} we observe that the PIM kernel time scales linearly with the PIM cores. On average, across all RL workloads for FP32 and INT32, the speedup from 125 PIM cores to 2,000 PIM cores exceeds $15\times$. The scaling for the taxi environment also follows a similar trend in performance scaling~(Figure~\ref{Figure6}). This speedup can be attributed to the large memory bandwidth and increased concurrency offered by scaling the number of PIM cores.

\kailash{Second}, the communication cost between PIM cores is relatively small for RL workloads with the frozen lake scenario due to minimal data transfer between PIM cores. The largest fraction of Inter PIM core synchronization over the total execution is $21.19\%$ for {\fontfamily{lmtt}\selectfont Q-learner-STR-INT32} with 2000 PIM cores for the taxi environment. Even so, 2000 PIM core configuration significantly reduces the overall execution time for {\fontfamily{lmtt}\selectfont Q-learner}. This trend is similar in {\fontfamily{lmtt}\selectfont Q-learner-SEQ-INT32} with
$20.79\%$ over the total execution time spent on inter-PIM core communication, followed by $17\%$ for {\fontfamily{lmtt}\selectfont Q-learner-RAN-INT32} in the taxi environment~(Figure~\ref{Figure6}). This is because the taxi environment requires approximately $47\times$ more data~(Q-values compared to frozen lake~\cite{brockman2016openai}) that needs to be \kailash{transferred} between the PIM cores. Overall, the prominence of inter-PIM core communication correlates with the amount of data exchanged, with taxi environment exhibiting dominance due to its higher data exchange demands compared to frozen lake. 

\kailash{Third}, in terms of initial setup costs, {\fontfamily{lmtt}\selectfont Q-learner-STR-INT32} incurs the highest cost at $29.61\%$ in the frozen lake environment due to the initial dataset transfer~(CPU-PIM) amount to each PIM core is comparatively more substantial than the small amount of data~(Q-values) being transferred between the PIM cores. 

\kailash{Fourth}, the communication cost associated with initial setup (CPU-PIM) and final PIM-CPU transfers exhibit negligible overhead on the total execution time for the taxi environment.


\vspace{-\baselineskip}
\subsection{Comparison with CPU/GPU platforms}
\label{comparison_cpu_gpu}

We implemented the CPU and GPU versions of all of our RL algorithms that are widely considered as state-of-the-art baselines~\cite{sutton2018reinforcement, watkins1992q, li2020sample, mnih2016asynchronous}: 
(1) CPU-V1: Multiple threads update a shared Q-table through the update function. Each thread operates on a portion of the dataset independently, and the same Q-table is used for updates, and (2) CPU-V2: Distributed version, where multiple threads update the portions of data independently by using local Q-tables. 



Figure~\ref{Figure7} illustrates the execution times of Q-learning and SARSA learning on PIM, CPU, and GPU with frozen lake and taxi environments~\cite{brockman2016openai}. \kailash{We make four observations. First, from our analysis, we} observe that both {\fontfamily{lmtt}\selectfont Q-learner} and {\fontfamily{lmtt}\selectfont SARSA} show higher execution times across SEQ/STR/RAN due to floating-point operations, which are not natively supported by our PIM architecture~\cite{UPMEM, UPMEM_1}. Despite this limitation, our {\fontfamily{lmtt}\selectfont Q-learner-SEQ-FP32} and {\fontfamily{lmtt}\selectfont SARSA-SEQ-FP32} workloads are $1.84\times$ and $2.08\times$ faster than their counterpart CPU versions~(CPU-V1) for frozen lake task. 

\kailash{Second}, for frozen lake, when using random sampling to prioritize exploration~({\fontfamily{lmtt}\selectfont Q-learner-RAN-FP32}), we observed a speedup of $1.96\times$ compared to the CPU-V1 version. However, in the taxi environment, compared to the CPU-V1, our PIM implementation~({\fontfamily{lmtt}\selectfont Q-learner-SEQ-RAN-STR-FP32}) is almost $0.64\times$ slower on average due to the large number of floating point operations performed corresponding to huge state-action size. Compared to the CPU-V2 in the taxi environment, we observe a slowdown in execution time for sequential and stride-based sampling techniques against the CPU-V2. This is due to CPU hardware prefetcher's ability to enhance cache locality for sequential and stride memory access patterns, where CPU-cache latencies are lower than that of PIM-DRAM. 

\kailash{Third}, using fixed-point representation (INT32) offers higher performance than the floating-point (FP32) format. For example, {\fontfamily{lmtt}\selectfont Q-learner-SEQ-INT32} is $8.16\times$ faster than {\fontfamily{lmtt}\selectfont Q-learner-SEQ-FP32}, and the trend is similar across various sampling strategies. This is the result of using natively supported instructions (even though 32-bit integer multiplications are emulated by the run-time library~\cite{gomez2022benchmarking, gomez2021benchmarking}). 

\kailash{Fourth}, the GPU version of Q-learning with sequential sampling outperforms {\fontfamily{lmtt}\selectfont Q-learner-SEQ-FP32-FL} by $1.68\times$, benefiting from the Ampere architectures' large set of SIMD lanes and enhanced memory bandwidth. Notably, our {\fontfamily{lmtt}\selectfont Q-learner-SEQ-INT32-FL} achieves a substantial speedup of $4.84\times$ over the GPU version due to INT32 instructions. The {\fontfamily{lmtt}\selectfont SARSA-SEQ-INT32-FL} achieves a speedup over {\fontfamily{lmtt}\selectfont SARSA-SEQ-FP32-FL} by $4.73\times$.



Finally, our findings highlight the UPMEM architectures potential for accelerating the training of multiple independent Q-learners, where each agent trains an offline dataset of size 10,000~(frozen lake) transitions and learns individual optimal policies. To illustrate, when training 1,000 agents, each with 10,000 transitions, for 2,000 episodes, the overall execution time is approximately 996.52 seconds. Scaling up to 2,000 agents increases the execution time to about 1,943.78 seconds on an Xeon CPU. 

Our PIM implementation with fixed-point representation introduces agent-level parallelism, demonstrating algorithmic scalability. By training individual agents on PIM cores, we achieve significant speedup compared to their baseline CPU version, which utilizes multiple independent Tabular Q-learners. Specifically, \name achieves a speedup of approximately $11.23\times$ for 1,000 agents and $21.92\times$ for 2,000 agents when executing on 1,000 and 2,000 PIM cores, respectively.

\section{Key Takeaways}
\label{Observations}
Our design and evaluation of \name, the first-known implementation that accelerates RL workloads on processing-in-memory systems, gave us several insights:
\begin{itemize}
\item RL workloads demonstrate reduced performance potential on UPMEM hardware PIM accelerator due to instruction emulation by the runtime library as the floating-point operations are not supported by the UPMEM platform. To tackle this, we adopt 32-bit fixed-point representations~(Section~\ref{Performance_analysis_PIM_kernel}).
\item The most suited reinforcement learning (RL) algorithms for the UPMEM PIM architecture are those that have memory-intensive tasks and require minimal communication between the inter-PIM cores. For instance, our study shows \kailash{multi-agent} Q-learning demonstrates better hardware adaptability~(Section~\ref{comparison_cpu_gpu}).
\item Scaling the PIM cores linearly leads to a nearly proportional reduction in the execution time  for a given working set size~(Section~\ref{Performance_analysis_PIM_kernel}). 
\item We demonstrate that PIM is beneficial for random memory accesses. However, when it comes to accessing data sequentially or in stride-based patterns, the CPU hardware prefetcher's strong capability in managing data locality results in improved performance~(Section~\ref{comparison_cpu_gpu}).
\end{itemize}

\section{Related Work}
\label{Related work}

\kailash{To our knowledge, our work represents the first instance of adapting RL on real \kailash{PIM architectures}. We have already extensively compared \name to state-of-the-art CPU-based and GPU-based systems and presented the strong scaling experiments in Sections~\ref{comparison_cpu_gpu} and \ref{Performance_analysis_PIM_kernel}, respectively. In this section, we briefly summarize other related works in two categories:~\circled{1} Outlining recent advancements in leveraging PIM systems to accelerate workloads, including deep learning and machine learning.~\circled{2}~Reviewing prior efforts to accelerate RL and highlighting how our work distinguishes itself.}

\subsection{Processing-in-Memory Systems}

\subsubsection{DL Training and Inference}

Prior efforts leverage PIM systems to accelerate deep learning~(DL) inference and training phases~\cite{das2022implementation, li2020hitm, schuiki2018scalable, luo2020benchmark, sun2020energy, liu2018processing, he2020newton} . For instance, various proposals have been studied to accelerate DL inference phases, including CNNs~\cite{deng2018dracc, boroumand2018google, gao2017tetris}, RNNs~\cite{boroumand2021google, lee20221ynm}, and recommendation systems~\cite{niu2022184qps, ke2021near}. Another avenue of exploration in academia and industry capitalizes on the analog computation capabilities of non-volatile memories (NVMs), particularly for tasks like matrix-vector multiplication, thereby facilitating the training of deep neural networks~\cite{luo2020benchmark, sun2020energy, imani2019floatpim}. Samsung's AxDIMM is an illustrative prototype, embedding an FPGA fabric in the DIMM's buffer chip, specifically designed to accelerate recommendation inference in Deep Learning Recommendation Models (DLRMs)~\cite{ke2021near}. Additionally, SK Hynix has introduced the Accelerator-in-Memory, a PIM architecture based on GDDR6, featuring specialized multiply-and-accumulate units and lookup-table-based activation functions to expedite deep learning workloads~\cite{9882182}.

\subsubsection{PIM for ML algorithms}
Few related prior works propose solutions for ML algorithms and evaluate the performance benefits of PIM technologies~\cite{gao2015practical, vieira2018exploiting, falahati2018origami}. For instance, UPMEM PIM is the first real-world processing-in-memory architecture used to accelerate ML workloads encompassing tasks like linear regression, logistic regression, and K-nearest neighbors~\cite{gomez2023evaluating}. Another line of work leverages different memory technologies~(e.g., 3D-stacked DRAM~\cite{gao2015practical, falahati2018origami}, SRAM~\cite{vieira2018exploiting}) to accelerate memory-bound machine learning applications~\cite{bavikadi2020review, 9739030, vieira2018exploiting, falahati2018origami}.
None of these works present a comprehensive implementation and evaluation of RL algorithms utilizing a real processing-in-memory architecture. 

\subsubsection{\kailash{UPMEM PIM system}}
\kailash{Several studies have focused on characterizing and outlining the architecture of UPMEM's PIM system~\cite{gomez2021benchmarking, gomez2022benchmarking,nider2021case,peccerillo2022survey,devaux2019true}. There are several works that explore accelerating variety of applications and algorithms UPMEM’s PIM system, such as ML training/inference~\cite{gomez2022benchmarking, gomez2021benchmarking,gomez2022machine, gomez2023evaluating,giannoula2024accelerating, das2022implementation,zarif2023offloading,kim2024optimal,wu2023pim}, bioinformatics~\cite{chen2023uppipe,diab2022high,diab2023framework,lavenier2020variant,lavenier2016blast}, analytics \& databases~\cite{bernhardt2023pimdb,lim2023design,baumstark2023adaptive,baumstark2023accelerating}, security~\cite{gupta2023evaluating,jonatan2024scalability}, distributed optimization algorithms~\cite{rhyner2024analysis} and more~\cite{giannoula2022towards, oliveira2023transpimlib, chen2023simplepim, abecassis2023gapim, diab2022high, giannoula2024accelerating, giannoula2022sparsep,oliveira2022accelerating, hyun2024pathfinding}. However, none of the prior works have explored reinforcement learning (RL) algorithms on UPMEM’s PIM system, a gap that we fill by implementing and conducting a comprehensive evaluation in this paper.}

\subsection{Accelerating Reinforcement Learning Workloads}

\subsubsection{Distributed Training}
Prior works on distributed training have been proposed to accelerate the training phase of RL workloads~\cite{babaeizadeh2016ga3c, cho2019fa3c, li2019accelerating, hoffman2020acme, stooke2018accelerated, clemente2017efficient}. Another strategy for multi-agent RL acceleration is to restrict the agent interactions to one-hop neighborhoods and adopt a distributed training strategy to accelerate the training phase~\cite{wang2022darl1n}. However, training on VM-based approaches still requires extensive management of the cluster and deploying the training jobs. Prior studies, like FA3C~\cite{cho2019fa3c}, have focused on accelerating multiple parallel worker scenarios, where each agent is controlled independently within their own environments using single-agent RL algorithms. Contrary to that, \name designs a distributed learning architecture, with PIM cores executed concurrently without extra cluster management.

\subsubsection{Quantization}
Low-precision~(Quantization) training for neural networks reduces the neural network weights, enables faster compute operations, and minimizes the memory transfer computation time. Quantization aware training~\cite{dong2019hawq, hubara2017quantized}, post-quantization training~\cite{tambe2020algorithm, zhao2019improving}, and mixed precision~\cite{micikevicius2017mixed} demonstrated that neural networks may be quantized to a lower precision without significant degradation in the accuracy or rewards. Furthermore, to speed up the training, prior works have shown that half-precision quantization can yield significant performance benefits and improve hardware efficiency by reducing precision from FP32 to FP16 or even lower while achieving adequate convergence~\cite{bjorck2021low}. Other relevant approaches include QuaRL~\cite{krishnan2022quarl}, where the authors demonstrated that applying quantization on RL algorithms and quantizing the policies down to ≤ 8 bits led to substantial speedups compared to full precision training. In contrast, we accelerate the training phase of offline RL workloads with large datasets on a real-world PIM architecture that exhibits a memory-bounded behavior. 


\section{Conclusion}
\label{Conclusion}

\kailash{In this paper, by adapting and implementing popular RL algorithms on a real Processing-in-Memory~(PIM) architecture\ms{,} we explore the potential of memory-centric systems in Reinforcement Learning~(RL) training. We evaluate our PIM-based Q-learning and SARSA algorithm implementations on the \kailash{UPMEM PIM} system with up to 2000 PIM cores. We explore} several optimization strategies that will enhance the performance of these RL workloads under different input data types and sampling strategies. We evaluate the quality, performance, and scalability of RL workloads on \kailash{PIM architectures} compared to state-of-the-art CPU and GPU baselines. Our findings indicate that PIM systems offer superior performance compared to CPUs and GPUs when handling memory-intensive RL workloads. Our studies demonstrate a near-linear scaling of 15× in performance when the number of PIM cores increases by a factor of 16× (125 to 2000). Our research results demonstrate that PIM systems have the potential to serve as \kailash{effective} accelerators for a diverse range of RL algorithms in the future.


\begin{acks}
This research is based on work supported by the National Science Foundation under grant CCF-2114415. \ms{We thank UPMEM for providing hardware resources to perform this research. We acknowledge the generous gifts from our industrial partners, including Google, Huawei, Intel, and Microsoft. This work is supported in part by the Semiconductor Research Corporation (SRC), the ETH Future Computing Laboratory (EFCL), and the AI Chip Center for Emerging Smart Systems (ACCESS).}

\end{acks}

\balance
{
  \bstctlcite{IEEEexample:BSTcontrol} 
   \let\OLDthebibliography\thebibliography
  \renewcommand\thebibliography[1]{
    \OLDthebibliography{#1}
    \setlength{\parskip}{0pt}
    \setlength{\itemsep}{0pt}
  }
  \bibliographystyle{IEEEtran}
  \bibliography{references}
}

\end{document}